# Multi-AD: Cross-Domain Unsupervised Anomaly Detection for Medical and Industrial Applications

Wahyu Rahmaniar[a,*] and Kenji Suzuki[a]

[a]*BioMedical Artificial Intelligence (BMAI) Research Unit, Institute of Integrated Research, Institute of Science Tokyo[*], Kanagawa 226-8503, Japan*

**Abstract**

*Traditional deep learning models often lack annotated data, especially in cross-domain applications such as anomaly detection, which is critical for early disease diagnosis in medicine and defect detection in industry. To address this challenge, we propose Multi-AD, a convolutional neural network (CNN) model for robust unsupervised anomaly detection across medical and industrial domain images. Our approach employs the squeeze-and-excitation (SE) block to enhance feature extraction via channel-wise attention, enabling the model to focus on the most relevant features and detect subtle anomalies. Knowledge distillation (KD) transfers informative features from the teacher to the student model, enabling effective learning of the differences between normal and anomalous data. Then, the discriminator network further enhances the model's capacity to distinguish between normal and anomalous data. At the inference stage, by integrating multi-scale features, the student model can detect anomalies of varying sizes. The teacher-student (T-S) architecture ensures consistent representation of high-dimensional features while adapting them to enhance anomaly detection. Multi-AD was evaluated on several medical datasets, including brain MRI, liver CT, and retina OCT, as well as industrial datasets, such as MVTec AD, demonstrating strong generalization across multiple domains. Experimental results demonstrated that our approach consistently outperformed state-of-the-art models, achieving the best average AUROC for both image-level (81.4% for medical and 99.6% for industrial) and pixel-level (97.0% for medical and 98.4% for industrial) tasks, making it effective for real-world applications.*

## 1. Introduction

Anomaly detection (AD) is a critical task in the medical and industrial domains, where early identification of anomalous patterns can help ensure patient safety, operational efficiency, and product quality. In healthcare, detecting anomalies such as tumors, lesions, and pathological

[*]Tokyo Institute of Technology and Tokyo Medical and Dental University have been merged to establish Institute of Science Tokyo on October 1, 2024.

*Corresponding author

*Email addresses:* rahmaniar.w.aa@m.titech.ac.jp (W. Rahmaniar); suzuki.k.di@m.titech.ac.jp (K. Suzuki)

structures is crucial for timely diagnosis and effective treatment, directly impacting patient outcomes and survival rates [1,2]. Similarly, identifying defects in manufacturing processes or equipment failures in industrial settings can help maintain high-quality production standards and prevent costly operational disruptions [3,4]. However, the scarcity of labeled anomaly data is a problem with AD, as such cases are typically rare, and acquiring labeled data, particularly in medical imaging, is labor-intensive, expensive, and subject to inter-observer variability [5-7]. These limitations are particularly pronounced in fields such as radiology, where subtle or heterogeneous anomalies necessitate large and diverse datasets that are rarely available [8].

To address the scarcity of labeled data, recent advances have turned to self-supervised learning (SSL) [9-11]. SSL techniques learn robust feature representations from unlabeled data by solving pretext tasks (*e.g.*, reconstructing masked regions), allowing the model to capture intrinsic data patterns without manual labeling. By training on "normal" data distributions, SSL-based AD frameworks can identify outliers that signal anomalies, offering a practical solution for domains where labeled anomalies are rare or non-existent [12]. However, although SSL reduces reliance on annotations, its performance suffers in scenarios with limited training data or high intra-class variability, such as medical imaging with diverse anatomical structures [13] or industrial systems operating in complex operational environments [14].

To bridge these gaps, knowledge distillation (KD) has emerged as a promising strategy to enhance AD robustness [15-17]. KD transfers knowledge from a larger "teacher" (*T*) model to a compact "student" (*S*) model, maintaining performance while increasing efficiency and generalization [18]. For instance, in medical imaging, KD enables lightweight models to mimic expert-level AD capabilities, even when trained on small datasets. However, existing KD approaches often focus on single-domain applications. They are seldom evaluated for cross-modality transfer, motivating the design of a system that separates domain-agnostic from domain-specific components [19,20].

Based on these advances, we propose Multi-AD, an unsupervised AD framework designed to address challenges in AD for medical and industrial applications. The key innovation of Multi-AD lies in its ability to generalize across multiple domains by leveraging domain-independent feature representation and domain-specific adaptation. Adopting optimized unsupervised learning and KD techniques, where features learned from a normal sample dataset are refined into an S model that effectively distinguishes between normal and anomalous features, even in the presence of limited labeled data. In summary, our contributions are as follows:

1) A cross-domain AD framework that disentangles domain-agnostic (shared) and domain-specific (adaptive) feature learning, improving generalization across medical and industrial datasets.
2) A convolution-enhanced multi-scale fusion module that improves the localization of small and large anomalies.
3) Discriminator (*D*) networks with adaptive attention mechanisms, enabling precise anomaly localization in various imaging modalities (*e.g.*, MRI, CT, and OCT) and industrial inspection systems
4) Provision of precise anomaly maps for accurate localization of diseases and product defects.

## 2. Related Works

Despite significant advances in AD for medical and industrial imaging, scalability, robustness, and cross-domain generalization remain challenging [21]. Traditionally, supervised learning approaches that rely on large-scale labeled datasets containing normal and defective samples have been used for AD [22, 23]. However, the scarcity of labeled anomalous samples and the wide variety of potential anomaly types have limited the effectiveness of these methods. The following section provides an overview of previous methods used in AD for medical and industrial applications.

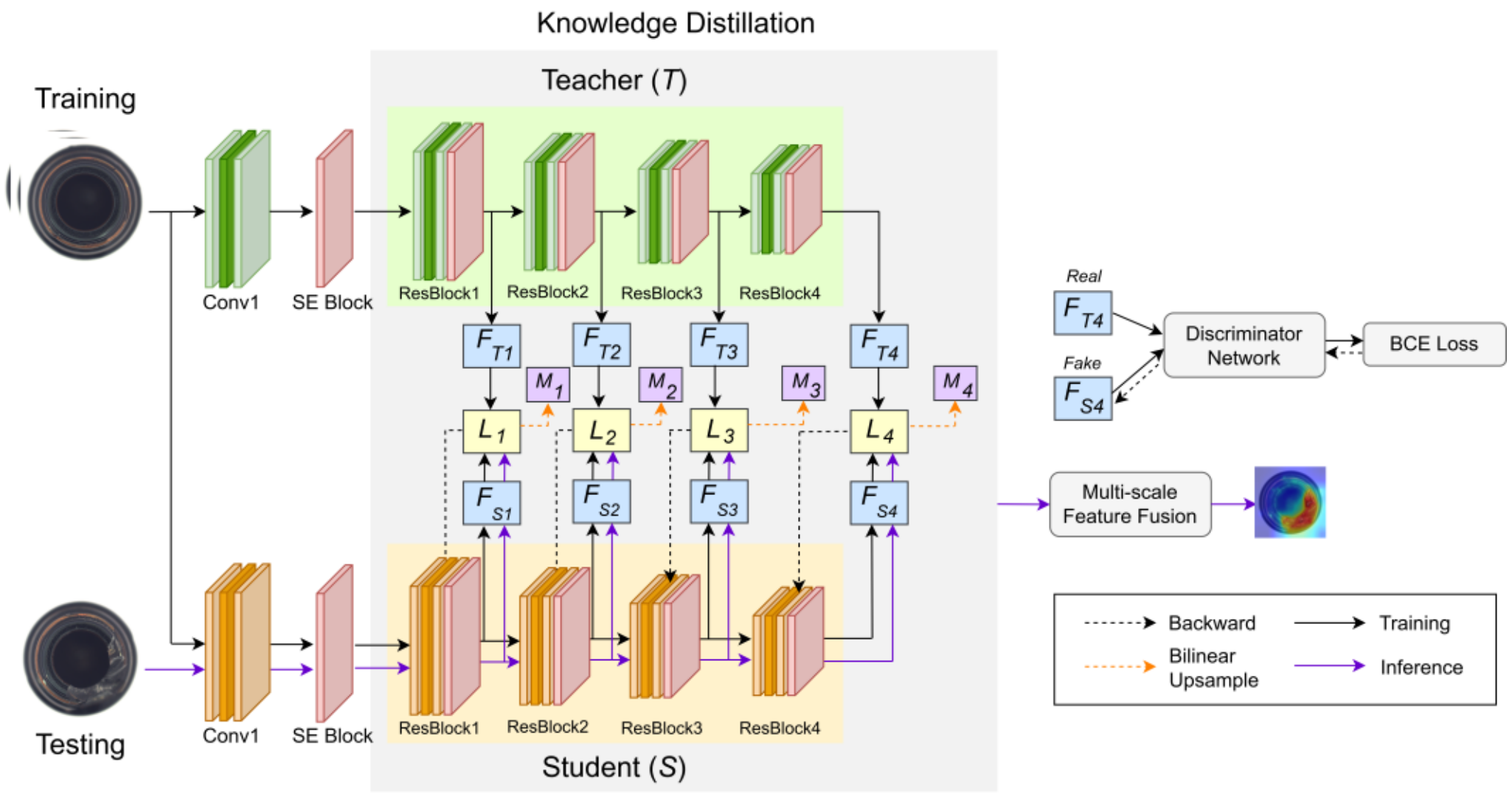

**Figure 1.** Proposed Multi-AD method for cross-domain anomaly detection.

*2.1 Anomaly Detection in Medical Images*

Generative adversarial networks (GANs) have been widely explored for AD in medical images, such as MADGAN for brain MRI [24]. The potential of GAN-based methods lies in their ability to generate high-quality images. However, they often encounter challenges, such as unstable training and mode collapse, that limit their scalability across medical imaging modalities [25].

Deep perceptual autoencoder (DPA) models can learn a compact representation of normal data in a lower-dimensional space, flagging outliers as anomalies based on reconstruction errors [26]. This approach has been applied to medical images, such as histopathology and chest X-rays, demonstrating improved robustness compared with traditional autoencoder methods [27]. On the other hand, its reliance on well-represented normal classes and the challenges of feature selection may limit its performance on highly complex medical datasets.

*2.2 Anomaly Detection in Industrial Images*

AD in industrial images helps maintain production standards, reduce waste, and minimize

**Algorithm 1** Training Procedure of Multi-AD

**Input:** Training dataset $X_t = \{X_1^t, X_2^t, \cdots, X_n^t\}$, pretrained *T* (teacher model) parameters $\beta_T$, epoch number *m*

**Output:** *S* (student model) parameters $\beta_S$;

1: Initialization:
- Randomly initialize $\beta_S$ and *D* (discriminator) parameter $\beta_D$;
- Load pretrained weight $\beta_T$ into *T* and freeze *T* (no gradient);

2: Feature extraction: For each input $X_j^t$ in $X_t$;

3: **for** *i* ← 1 to *m* **do**

4: **for** *j* ← 1 to *n* **do**

5: **Forward pass:** For each $X_j^t$ compute multi-layer features:

$$F_T \leftarrow T(X_j^t) \text{ and } F_S \leftarrow S(X_j^t);$$

6: **Feature normalization**: Apply L2 normalization to *T* or *S* feature maps used for distillation: $\hat{F}_T = {F_T}/{\|F_T\|_2}$ and $\hat{F}_S = {F_S}/{\|F_S\|_2}$;

7: **Loss computation:**
- Feature alignment (distillation): Compute the feature generator loss $\mathcal{L}_G$ by comparing feature activations across critical layers on Eq (8);
- Discriminator loss: Compute $\mathcal{L}_D$ using real (*T*) and fake (*S*) labels on Eq (10);
- *S* adversarial term: Compute $\mathcal{L}_{adv}$ Eq (11);
- *S* total loss: $\mathcal{L}_S$ on Eq (12), integrating both feature generator and discriminator losses;

8: **Parameter update:**
- Update *D:* $\beta_D \leftarrow \beta_D - \eta_D \nabla_{\beta_D} \mathcal{L}_D$ (freeze *S*, *T* frozen) ;
- Update *S*: $\beta_S \leftarrow \beta_S - \eta_S \nabla_{\beta_S} \mathcal{L}_S$ (freeze *D*, *T* frozen) ;

9: **end for** *j*

10: **end for** *i*

operational costs. However, challenges such as limited labeled data, variability in defect types, and the need for real-time processing make this task complex [28,29].

Feature-matching-based methods use a pre-trained feature extractor to extract features from anomaly-free samples and build a feature memory bank [30,31]. These methods match query image features with database features during inference to identify anomalies, but they incur significant memory overhead due to the extensive storage required for feature representations. They can also incur higher computational costs when down-sampling features to manage memory [32]. Feature matching operates at the patch level and treats samples as unordered local feature sets; these approaches often miss positional details and fail to detect anomalies that require spatial context or occur at specific locations [33].

Unsupervised AD methods have been widely explored in industrial settings because normal data are more abundant than defective examples [34, 35]. Although GANs for AD have shown promising results on industrial datasets, they often suffer from mode collapse and difficulty reproducing high-fidelity reconstructions, particularly in scenarios involving complex or irregular defects [36]. Moreover, the prior normality prompt transformer (PNPT) has been successfully applied to various industrial datasets, including texture, object, and surface data [37]. However, the transformer's reliance on well-represented normal classes and its complexity in managing a dual-stream architecture can make it sensitive to feature selection and computationally intensive, thereby limiting its scalability across diverse industrial imaging tasks [38].

## 3. Proposed Method

Multi-AD integrates unsupervised learning and KD via domain-agnostic feature learning to capture universal anomalous patterns (*e.g.*, irregular textures or shapes) and domain-specific adaptation, thereby refining the representation for the target modality, as illustrated in Figure 1. This allows the framework to generalize across unseen domains while maintaining fine-grained anomaly sensitivity. By distilling features learned from the *T* model (trained on normal samples) into the *S* model (tested on both normal and anomalous data), the model's discriminative ability can be improved even with minimal labels. The specialized discriminator improves localization accuracy across multiple modalities. Furthermore, the proposed multi-scale feature extraction mechanism enables the simultaneous detection of global anomalies (*e.g.*, large lesions) and local defects (*e.g.*, microcracks in machines). This capability is critical in real-world scenarios where anomalies manifest at varying scales and intensities.

Algorithm 1 describes the overall training process of the proposed Multi-AD framework, which integrates teacher-student (*T-S*) KD, discriminator (*D*) optimization, and multi-scale anomaly localization. The algorithm outlines the main steps in the subsections, including feature extraction, loss function formulation, and update strategies for the *S* and *D* networks. This structure ensures that each component contributes cohesively to the final anomaly map, rather than functioning in isolation.

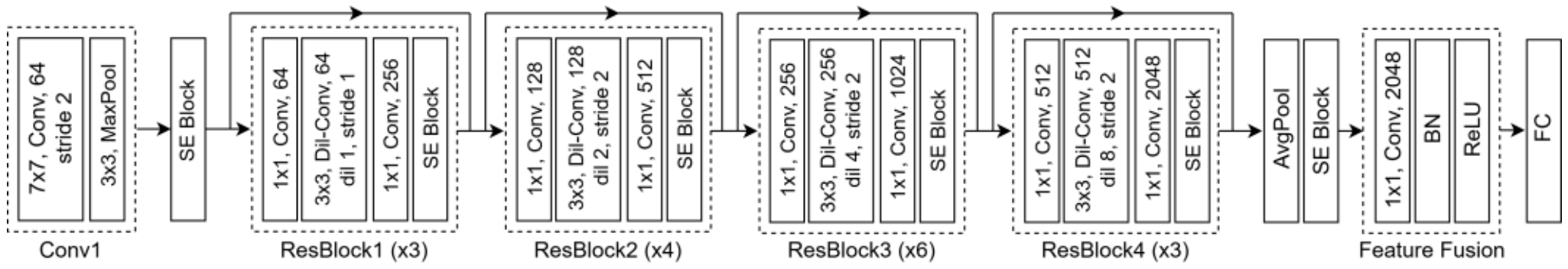

**Figure 2.** Proposed backbone network.

**Table 1.** Comparison of the original WideResNet backbone with the improved version.

| Feature | Original WideResNet | Improved WideResNet |
|---|---|---|
| Convolutional layers | Standard convolutions | Dilated convolutions for expanded receptive fields |
| Feature recalibration | Not present | SE blocks added to recalibrate channel-wise feature responses |
| Feature fusion | Absence of explicit fusion | Convolutional layer for multi-scale feature fusion |
| Focus on multi-scale context | Limited, relies on depth and pooling | Enhanced through dilated convolutions and fusion techniques |
| Receptive field | Standard size (depends on layer depth) | Expanded due to dilated convolutions |
| Scalability and flexibility | Good scalability, less focus on feature adaptivity | Improved scalability and adaptability to feature variations |

### *3.1 Backbone*

Figure 2 illustrates the proposed backbone network, a modification of the WideResNet-50 architecture [39] that integrates squeeze-and-excitation (SE) blocks [40] to improve AD performance across medical and industrial domains. The backbone architecture consists of an initial 7×7 convolutional layer with 64 filters, a stride of 2, and a padding of 3, followed by batch normalization (BN), a rectified linear unit (ReLU) activation, and a 3×3 max-pooling operation. The choice of a large initial convolutional kernel (7×7) ensures that spatial information is preserved across a sizeable receptive field at the beginning of the network. The proposed backbone is based on an architecture that balances depth and computational efficiency. The number of feature planes per layer can be increased, enabling richer feature representations without a significant increase in parameters.

Dilated convolutions are applied at the four residual stages (ResBlock1, ResBlock2, ResBlock3, ResBlock4) to enlarge the receptive field while maintaining the spatial resolution. For a convolution with a dilation rate *r*, the output is computed as:

$$y[i, j] = \sum_m \sum_n x[i + r \cdot m, j + r \cdot n]\omega[m, n], \quad (1)$$

where $x$ denotes the input feature map, $\omega[m, n]$ represents the convolutional kernel at index *(m, n),* and $y$ denotes the resulting output. The dilation rate is progressively increased across the residual stages ($r$ = 1, 2, 4, 8) to enable the network to capture both local details and global context.

This design enables the model to detect minor, localized anomalies (*e.g.*, subtle lesions) and diffuse patterns (*e.g.*, significant structural distortions) in the images. Such versatility is important for addressing anomalies that vary in shape, size, and location. By using dilated convolutions, the model avoids resolution loss typically associated with down-sampling layers, which is critical for preserving spatial detail. Each ResBlock in the network architecture incorporates BN after each convolution to normalize the output and stabilize the learning process. This is followed by a ReLU activation function, which introduces non-linearity and increases the network's capacity to learn complex features. Additionally, each block includes a shortcut connection that combines its input with its output, mitigating the vanishing gradient problem by promoting more effective gradient flow during training.

The SE blocks are integrated into each residual stage to recalibrate channel-wise feature responses, allowing the network to emphasize informative features while suppressing less relevant ones. The process begins with a squeeze operation that employs global average pooling to compute channel-wise statistics. For each channel in $c$ in an input feature map of $x$ of dimensions $h \times w$, the statistic is calculated as:

$$z_c = \frac{1}{h \times w} \sum_{i=1}^{h} \sum_{j=1}^{w} x_c(i, j), \quad (2)$$

where $x_c(i, j)$ represents the activation at a spatial location $(i, j)$ for channel $c$. Next, the pooled vector $z_c$ is processed through two fully connected (FC) layers. A ReLU activation follows the first

FC layer, and the second applies a sigmoid function to yield a channel-wise scaling factor by:

$$s_c = \sigma\big(\tau_2 . \mathrm{ReLU}(\tau_1 \cdot z_c)\big), \tag{3}$$

where $\tau_1$ and $\tau_2$ are learnable weight matrices and $\sigma$ denotes the sigmoid activation function. Finally, the recalibration is performed by scaling each channel $c$ of the original feature map by its corresponding factor $s_c$ by:

$$\hat{x}_c(i,j) = s_c \cdot x_c(i,j), \tag{4}$$

This recalibration enhances the network's ability to identify subtle anomalies by accentuating channels with more informative features and diminishing irrelevant channels. Including SE blocks further enables the network to adaptively focus on channel-specific cues relevant to the anomalous context of each dataset, thereby enhancing the domain adaptability of the proposed framework.

Furthermore, feature fusion integrates fine details from lower layers with the high-level semantic information extracted by upper layers. This fusion is implemented by applying 1×1 convolution followed by BN and ReLU activations, which can be expressed as:

$$F_{fused} = \mathrm{ReLU}\Big(\mathrm{BN}\big(\omega * [F_{low}, F_{up}]\big)\Big). \tag{5}$$

where $\omega$ represents the 1×1 convolution kernel and $[F_{low}, F_{up}]$ denotes the concatenation of the lower-level and the upper-level feature maps.

Table 1 summarizes the backbone comparison of the original WideResNet with the improved version. Compared with the original WideResNet, the improved version not only widens the receptive field and introduces dynamic feature recalibration but also enhances the network's scalability and flexibility in handling feature variations and patterns. This results in a more adaptive approach to handling diverse and complex datasets, which is especially important in precise anomaly detection across scales and contexts.

*3.2 Knowledge Distillation (KD)*

The KD process enables the *S* model to imitate the *T* model's feature representations by transferring mid-level knowledge. The *S* model achieves state-of-the-art performance by matching the hierarchical features extracted from the pre-trained *T* model, thereby effectively distinguishing normal from abnormal regions during testing. This intermediate-level distillation balances low-level texture transfer and high-level semantic abstraction, resulting in stronger generalization to unseen anomalies.

The model is trained on a dataset of anomaly-free or normal images, denoted as $X_t = \{X_1^t, X_2^t, \cdots, X_n^t\}$, where each image $X_n \in \mathbb{R}^{w \times h \times c}$ has dimensions *w*, *h,* and *c* representing the image's width, height, and channel size, respectively. During training, the model learns to recognize and localize features that follow the same distributions as the normal training data. This enables it to detect deviations (anomalies) in samples drawn from a different distribution during testing.

To mitigate the bias present in pre-trained networks (often trained on unrelated datasets), the backbone processes features at multiple levels of abstraction. For example, the top layer of the network $(F_{T1}, F_{S1})$ produces high-resolution, fine-grained features encoding details, such as texture, edges, and color $(M_1)$. In contrast, the deeper layer $(F_{T4}, F_{S4})$ generates low-resolution, high-level features containing contextual information, such as shape and structure $(M_4)$.

Let $F_T^N$ and $F_S^N$ denote the feature maps of the *T* and *S* models at the *N*-th intermediate layer. To ensure consistency during training, the feature maps from both models are first normalized using L2 normalization as:

$$\hat{F}_T^N = \frac{F_T^N}{\|F_T^N\|_2} \text{ and } \hat{F}_S^N = \frac{F_S^N}{\|F_S^N\|_2}, \quad (6)$$

The feature generator (*G*) extracts activation vectors from the normalized feature maps. The activations extracted from a critical layer (*C*) of the *T* and *S* models are represented as $\phi_T^C$ and $\phi_S^C$, respectively. Each activation captures both magnitude and direction (spatial) information, preserving the intrinsic value and structure of the features. For a given critical layer *i* and for each spatial location or neuron *j*, the activation vector can be expressed as:

$$\phi_T^j(i) = G_j\left(\hat{F}_T^N\right) \text{ and } \phi_S^j(i) = G_j\left(\hat{F}_S^N\right), \tag{7}$$

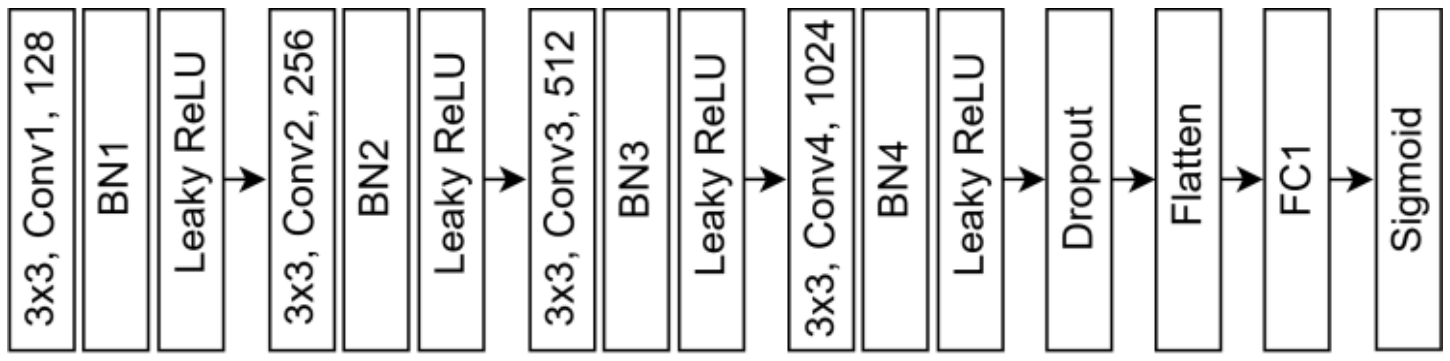

**Figure 3.** Proposed discriminator (*D*) network.

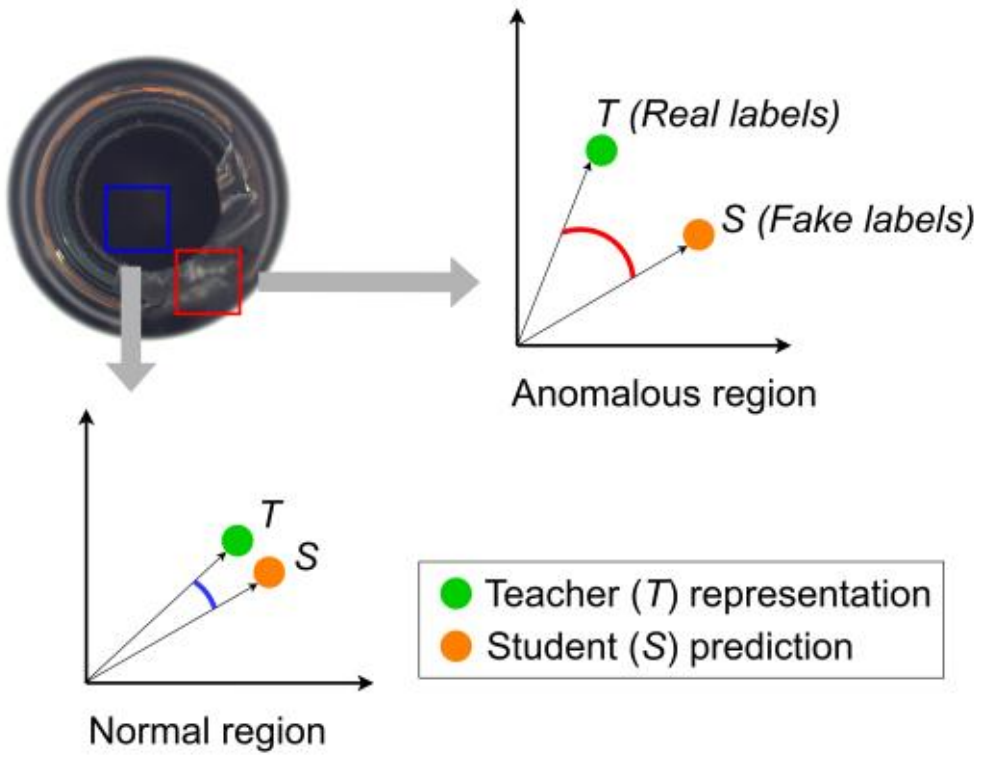

**Figure 4.** The student (*S*) model produces feature maps that are very similar to those of the teacher (*T*) model, enabling it to distinguish normal and anomalous regions effectively.

Knowledge transfer is achieved by aligning these normalized activations while minimizing the Euclidean distance. This alignment is enforced via the loss function defined as:

$$\mathcal{L}_G = \frac{1}{P}\sum_{i=1}^{P}\frac{1}{Q_i}\sum_{j=1}^{Q_i}\left(1 - cos\left(\phi_T^{(i,j)}, \phi_S^{(i,j)}\right)\right). \tag{8}$$

where $P$ is the total number of critical layers and $Q_i$ denotes the number of neurons in the $i$-th critical layer. Unlike traditional KD methods that focus solely on logits or final outputs, this approach leverages deep feature-level alignment, enabling more granular and spatially aware knowledge transfer. This loss ensures that the $T$ model's normalized activations effectively guide the $S$ model, facilitating robust knowledge transfer across the network layers.

*3.3 Discriminator (D) Network*

The *D* network comprises convolutional layers, followed by BN and a leaky ReLU activation, to progressively refine the input feature map, as illustrated in Figure 3. The architecture is structured to capture spatial hierarchies and complex patterns effectively. The first convolutional layer applies 128 filters with a kernel size of 3×3, a stride of 1, and a padding of 1, which preserves spatial resolution while extracting low-level features, defined as:

$$D_1(F^{(0)}) = \text{LeakyReLU}\left(\text{BN}\left(Conv_{3\times3}\left(F^{(0)}\right)\right)\right), \tag{9}$$

where $F^{(0)}$ is the input feature map. The subsequent layers further enhance the representation: the second convolutional layer uses 256 filters to refine mid-level features, the third employs 512 filters to capture high-level spatial relationships, and the fourth applies 1024 filters to generate a detailed representation. Then, a dropout layer is applied to reduce overfitting and improve generalization.

The output of the final convolutional layer is flattened into a one-dimensional vector and passed through the FC layer, reducing the dimensionality to a single scalar. A sigmoid activation function is applied to produce a probability score ranging from 0 to 1. This score indicates the likelihood of the input feature map being real (closer to the *T* model*)* or fake (closer to the *S* model). The *D* is optimized using a binary cross-entropy (BCE) loss function by

$$\mathcal{L}_D = -\frac{1}{m}\sum_{i=1}^{m}\left[\log D\left(F_T^{(i)}\right) + \log\left(1 - D\left(F_S^{(i)}\right)\right)\right], \tag{10}$$

where $D\left(F_T^{(i)}\right)$ and $D\left(F_S^{(i)}\right)$ are the *D*'s output for the *T* and *S* feature maps, respectively, and *m* is the batch size. This loss ensures that *D* assigns high probabilities to real features and low probabilities to fake features. Thus, this output provides a reliable estimate of how well the *S* model replicates the *T* model's internal representations, which serve as a dynamic feedback mechanism during training.

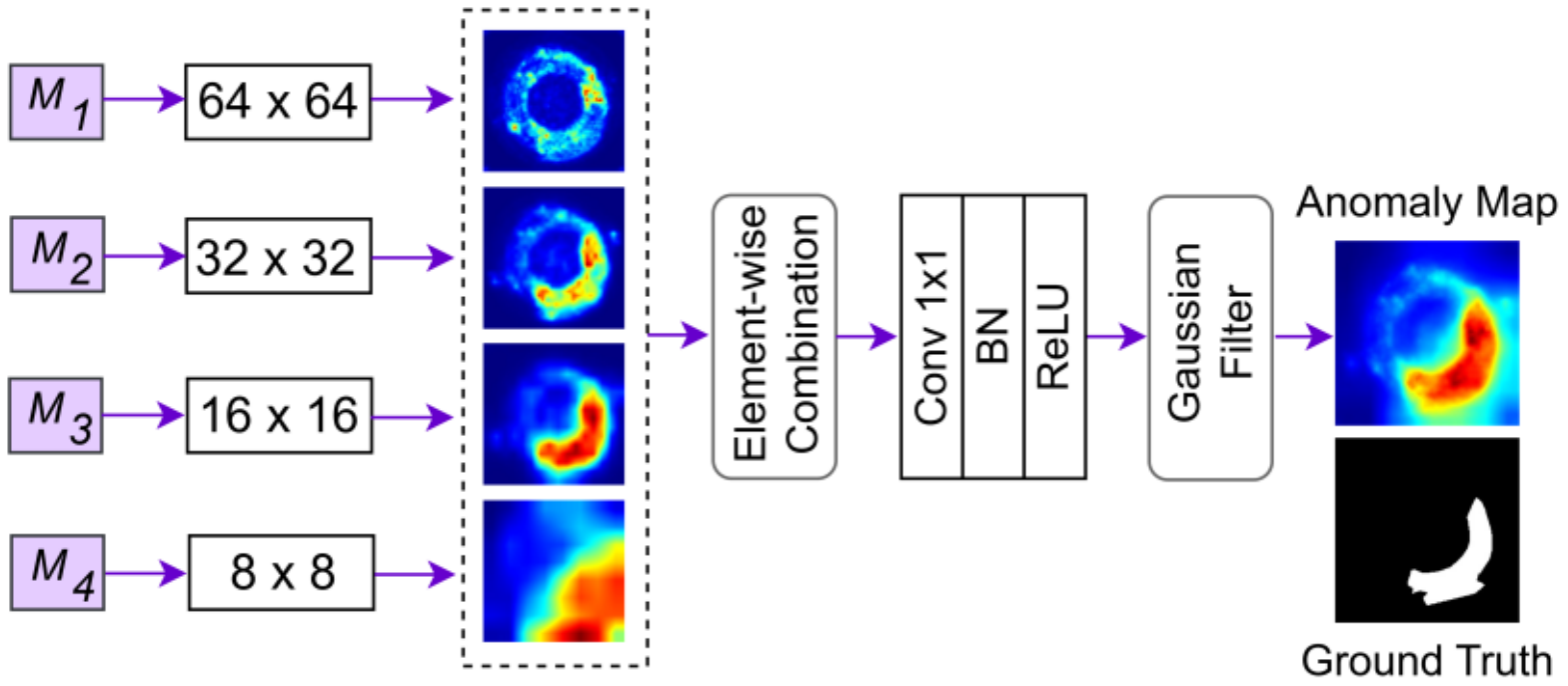

**Figure 5.** Convolution-enhanced multi-scale feature fusion.

During adversarial training, *D* learns to maximize its ability to distinguish genuine features (*T*) from fake features (*S*). In contrast, the *S* model seeks to minimize these differences, making its resulting features indistinguishable from those of the *T* model. These adversarial dynamics force the *S* model to produce feature maps that closely resemble those of the *T* model, thereby effectively capturing the data's normal distribution, as shown in Figure 4.

The optimization of the *S* model involves a combination of the feature-alignment loss $\mathcal{L}_G$ and an adversarial alignment term that encourages the student to fool *D* as follows:

$$\mathcal{L}_{adv} = -\frac{1}{m}\sum_{i=1}^{m} \log D\left(F_S^{(i)}\right), \tag{11}$$

The *S* model total loss is defined as:

$$\mathcal{L}_S = \mathcal{L}_G + \lambda \mathcal{L}_{adv}. \tag{12}$$

where $\lambda$ is a hyperparameter that controls the relative contribution of the two loss components. This total loss ensures a balance between feature alignment and adversarial discrimination, optimizing both aspects during training. Therefore, this combined optimization scheme enhances the fidelity of the *S*-model features.

*3.4 Convolution-Enhanced Multi-scale Feature Fusion*

To improve the robustness and accuracy of AD, we employ a multi-scale feature fusion (MFF) approach that integrates feature differences across multiple layers of the *T* and *S* models. Traditional MFF methods typically fuse multi-scale anomaly maps by simple summation, ensuring that feature representations from different network depths contribute equally to the final anomaly decision. However, these approaches often lack a refinement mechanism, leading to inconsistencies arising from differences in scale, feature distributions, and spatial misalignment across feature maps. To address these challenges, we introduce an enhanced MFF framework that incorporates convolutional operations to refine feature maps prior to fusion, thereby yielding a more robust and stable anomaly representation, as illustrated in Figure 5. This enhancement allows the model to adaptively calibrate the contributions of each scale before combining them, reducing overemphasis or underemphasis of certain features.

During inference, we compute pixel-wise cosine dissimilarity between normalized *T*- and *S*-features across multiple layers. The cosine loss function is the primary metric for detecting feature-level anomalies, as it quantifies the cosine difference at each pixel within the feature map.

Since anomalies can occur at different levels of abstraction, the model constructs multiscale anomaly maps to capture deviations across different spatial representations. These maps, $M_1$, $M_2$, $M_3$, and $M_4$, correspond to different depths in the network, each encoding unique structural characteristics. $M_1$ represents fine-grained texture variations typically observed in shallow layers, enabling the identification of small-scale anomalies. $M_2$ captures local structural irregularities as additional contextual information, indicating inconsistencies in feature distributions. $M_3$ encodes high-level spatial relationships, enabling the detection of broader semantic anomalies that may not be apparent at lower layers. Finally, $M_4$ incorporates global-scale feature irregularities, focusing on significant structural inconsistencies that indicate major anomalies across the image. By leveraging these hierarchical features, the model enhances its sensitivity and specificity, thereby reducing the likelihood of missing subtle anomalies or misclassifying normal patterns.

To localize anomalies at each stage of feature extraction, the pixel-wise cosine dissimilarity is computed for every spatial location in the feature map as follows:

$$M_{T-S}^{(n)}(i,j) = 1 - \frac{\sum_c \hat{F}_T^{(n,i,j,c)} \cdot \hat{F}_S^{(n,i,j,c)}}{\left\|\hat{F}_T^{(n,i,j)}\right\| \cdot \left\|\hat{F}_S^{(n,i,j)}\right\|}, \tag{13}$$

where $M_{T-S}^{(n)}(i,j)$ represents the pixel-level anomaly map computed at layer $n$, spatial position $(i,j)$, and channel $c$.

The proposed convolution-enhanced MFF incorporates an additional convolutional refinement step prior to fusion, which addresses inconsistencies arising from scale variations and misaligned feature maps. A 1×1 convolution layer reduces dimensionality while preserving spatial structure, ensuring that all layers contribute proportionally to the final representation of anomalies. BN stabilizes the feature distribution across layers, improving robustness by normalizing variations in feature responses. Finally, ReLU activation introduces non-linearity, enhancing the model's ability to capture complex patterns in anomalous regions. This additional processing step minimizes feature inconsistencies across scales, making the final anomaly map more robust. The anomaly maps are then combined via bilinear interpolation:

$$M_{fused} = \sum_{n=1}^{N} \Psi\left(M_{T-S}^{(n)}\right), \tag{14}$$

Followed by a Gaussian $\mathcal{G}_\theta$ smoothing operation to improve the clarity of detected anomalies and suppress high-frequency noise. The final image-level anomaly score is calculated by selecting the maximum response from the smoothed anomaly map:

$$\varphi = max\left(\mathcal{G}_\theta\left(M_{fused}\right)\right). \tag{15}$$

The proposed MFF enhances robustness by mitigating the limitations of single-layer feature maps, thereby combining complementary information from different network depths. Furthermore, this mechanism is inherently scalable, adapting smoothly to various anomaly sizes and distributions, making it suitable for multiple detection tasks. In practical applications, this flexibility enables Multi-AD to perform consistently across multiple imaging modalities and hardware configurations, eliminating the need for manual tuning or architectural changes.

## 4. Experimental Results

Model training and evaluation were performed on an RTX A6000 GPU and an Intel Core i9 CPU. The *T* model parameters were obtained from the backbone that was previously trained on ImageNet. In contrast, the *S* model was trained via distillation, leveraging knowledge transferred from the *T* model. Training focused solely on normal data; no anomalous images were encountered during training, whereas AD testing was applied to both normal and abnormal samples. All inputs were finally resized to 256 × 256 pixels. Optimization was performed using the Adam optimizer on the backbone and *D* networks, with a learning rate of 0.001 at 250 epochs and a batch size of 16.

### *4.1 Dataset*

a) Medical

To ensure a fair and representative evaluation of medical AD, we evaluate on three medical datasets (brain MRI, retinal OCT, and liver CT) commonly used in medical imaging AD [1]. This dataset encompasses a diverse range of imaging modalities and clinical scenarios, thereby providing a heterogeneous and clinically relevant testing platform. Each dataset represents a distinct anatomical and diagnostic context with unique imaging characteristics, thereby enhancing the robustness and generalizability of the evaluation.

The BraTS2021 [41] dataset consisted of 3D volumes transformed into 2D slices by extracting brain scans and corresponding annotations along the axial plane. We retained brain slices containing parenchyma (non-empty mask) and discarded empty background slices. This dataset comprises 11,298 slices, of which 7,500 normal images were used for training, with all images having a resolution of 240 × 240.

**Table 2.** Comparison of image-level ($I_L$) and pixel-level ($P_L$) AUROC (%) with state-of-the-art models on medical datasets.

| No | Dataset | Clustering-based | | | | | | Reconstruction-based | | | | Distillation-based | | | |
|---|---|---|---|---|---|---|---|---|---|---|---|---|---|---|---|
| | | PaDiM | | PatchCore | | SimpleNet | | STFPM | | RD4AD | | MKD | | Multi-AD (Ours) | |
| | | $I_L$ | $P_L$ | $I_L$ | $P_L$ | $I_L$ | $P_L$ | $I_L$ | $P_L$ | $I_L$ | $P_L$ | $I_L$ | $P_L$ | $I_L$ | $P_L$ |
| 1 | Brain | 79.0 | 94.4 | **91.6** | 96.7 | 77.7 | 93.7 | 83.0 | 95.6 | 89.5 | 96.4 | 81.5 | 89.4 | 89.7 | **96.8** |
| 2 | Liver | 50.8 | 90.9 | 60.3 | 96.4 | 52.5 | 96.8 | 61.8 | 91.2 | 60.4 | 96.0 | 60.7 | 96.1 | **62.6** | **97.6** |
| 3 | Retina | 75.8 | 91.4 | 91.6 | 96.5 | 87.3 | 94.7 | 84.8 | 91.2 | 87.7 | 96.2 | 89.0 | 86.7 | **91.8** | **96.7** |
| *Average* | | *68.5* | *92.2* | *81.2* | *96.5* | *72.5* | *95.1* | *76.5* | *92.7* | *79.2* | *96.2* | *77.1* | *90.7* | ***81.4*** | ***97.0*** |

**Table 3.** Comparison of image-level ($I_L$) and pixel-level ($P_L$) AUROC (%) with state-of-the-art models on industrial datasets.

| Dataset | | Clustering-based | | | | | | Reconstruction-based | | | | Distillation-based | | | |
|---|---|---|---|---|---|---|---|---|---|---|---|---|---|---|---|
| | | PaDiM | | PatchCore | | SimpleNet | | STFPM | | RD4AD | | MKD | | Multi-AD (Ours) | |
| | | $I_L$ | $P_L$ | $I_L$ | $P_L$ | $I_L$ | $P_L$ | $I_L$ | $P_L$ | $I_L$ | $P_L$ | $I_L$ | $P_L$ | $I_L$ | $P_L$ |
| Textures | Carpet | 96.2 | **99.1** | 98.7 | 99.0 | 99.7 | 98.2 | 98.8 | 95.8 | 98.9 | 98.9 | 79.3 | 95.6 | **99.8** | 98.9 |
| | Grid | 94.6 | 97.3 | 98.2 | 98.7 | **99.7** | 98.8 | 99.0 | 96.6 | 99.3 | **99.3** | 78.0 | 91.7 | **99.7** | 99.1 |
| | Leather | 97.8 | 99.2 | **100** | 99.3 | **100** | 99.2 | 99.3 | 98.0 | 99.4 | 99.1 | 95.1 | 98.0 | **100** | **99.4** |
| | Tile | 86.0 | 94.1 | 98.7 | 95.6 | **99.8** | 97.0 | 97.4 | 92.1 | 99.3 | 95.6 | 91.6 | 82.7 | **99.8** | **97.3** |
| | Wood | 91.1 | 94.9 | 99.2 | 95.0 | **100** | 94.5 | 97.2 | 93.6 | 99.2 | 95.3 | 94.3 | 84.8 | **100** | **96.9** |
| Objects | Bottle | 94.8 | 98.3 | **100** | 98.6 | **100** | 98.8 | 98.9 | 95.1 | 98.7 | 98.7 | 99.4 | 96.3 | **100** | **99.1** |
| | Cable | 88.8 | 96.7 | 99.5 | 98.4 | **99.9** | 97.6 | 95.5 | 87.7 | 95.0 | 97.4 | 89.2 | 82.4 | 99.6 | **98.5** |
| | Capsule | 93.5 | 98.5 | 98.1 | 98.8 | 97.2 | **98.9** | 98.3 | 92.2 | 96.3 | 98.7 | 80.5 | 95.9 | **98.4** | **98.9** |
| | Hazelnut | 92.6 | 98.2 | **100** | **98.7** | **100** | 97.9 | 98.5 | 94.3 | 98.9 | 98.6 | 98.4 | 94.6 | 99.8 | 98.2 |
| | Metal Nut | 85.6 | 97.2 | **100** | 98.4 | **100** | 98.8 | 97.6 | 94.5 | **100** | 97.3 | 73.6 | 86.4 | **100** | **98.9** |
| | Pill | 92.7 | 95.7 | 96.6 | 97.1 | 99.0 | 98.6 | 97.8 | 96.5 | 96.6 | 98.2 | 82.7 | 89.6 | **99.3** | **98.9** |
| | Screw | 94.4 | 98.5 | 98.1 | 99.4 | 98.2 | 99.3 | 98.3 | 93.0 | 97.0 | **99.6** | 83.3 | 96.0 | **98.6** | 99.3 |
| | Toothbrush | 93.1 | 98.8 | **100** | 98.7 | 99.7 | 98.5 | 98.9 | 92.2 | 99.5 | **99.1** | 92.2 | 96.1 | 99.8 | 98.9 |
| | Transistor | 84.5 | 97.5 | **100** | 96.3 | **100** | **97.6** | 82.5 | 69.5 | 96.7 | 92.5 | 85.6 | 76.5 | 98.9 | 95.1 |
| | Zipper | 95.9 | 98.5 | 98.8 | 98.8 | **99.9** | **98.9** | 98.5 | 95.2 | 98.5 | 98.2 | 93.2 | 93.9 | 99.8 | **98.9** |
| *Average* | | *92.1* | *97.5* | *99.1* | *98.1* | *99.5* | *98.2* | *97.1* | *92.4* | *98.2* | *97.8* | *87.8* | *90.7* | ***99.6*** | ***98.4*** |

The Retinal Edema Segmentation Challenge (RESC) [42] dataset comprises Optical Coherence Tomography (OCT) images of the retina, specifically targeting retinal edema cases. Each case includes 128 slices where retinal edema is present. The dataset comprises 6,217 images at 512 × 1024 resolution, of which 4,297 are normal images used for training.

The "Multi-Atlas Labeling Beyond the Cranial Vault" (BTCV) and Liver Tumor Segmentation (LiTs) datasets [43] comprise 3,201 slices at 512 × 512 resolution, including 1,542 normal slices, which were used for training. Hounsfield units (HU) from the 3D scans were converted to grayscale and cropped into 2D axial slices for analysis.

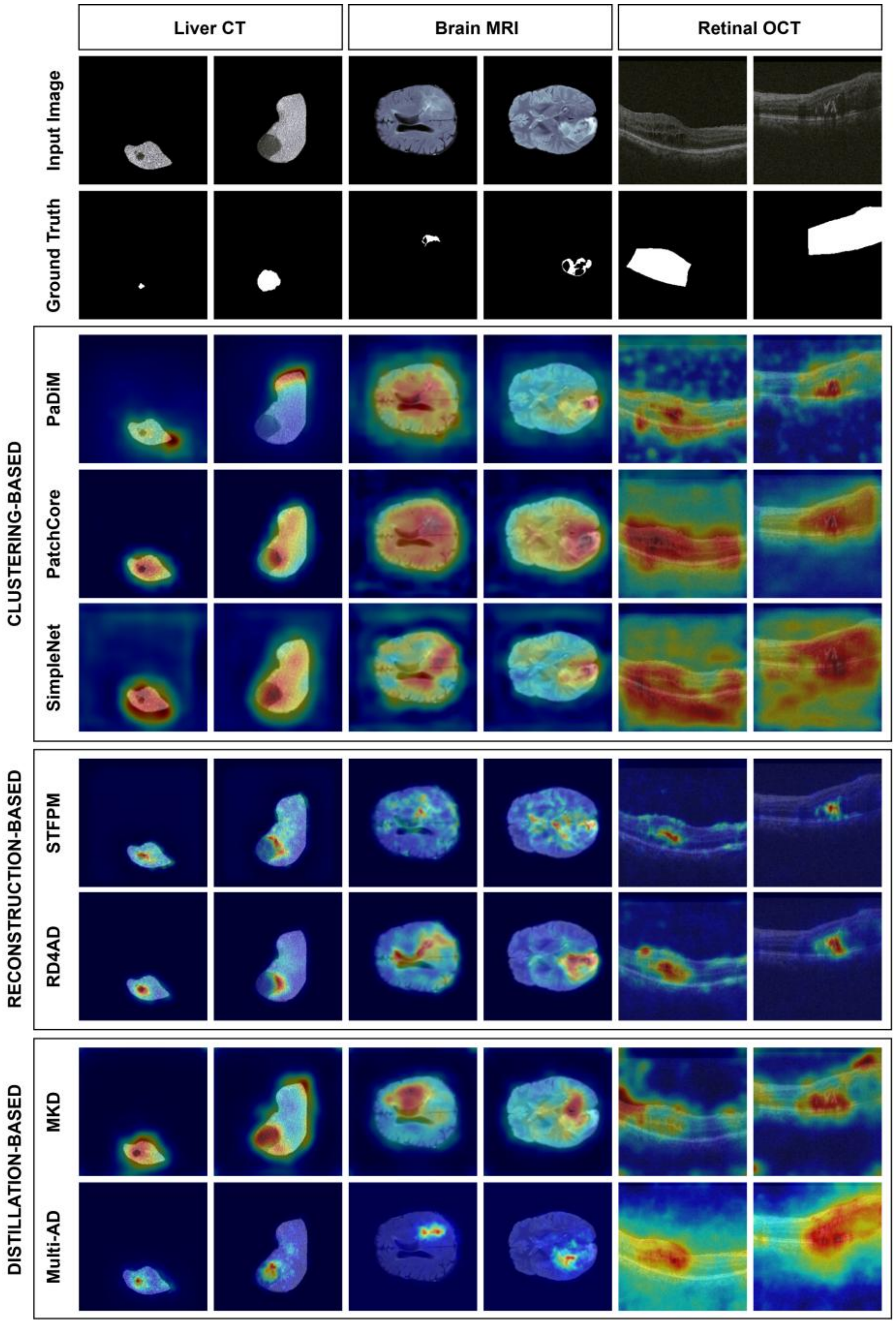

**Figure 6.** Anomaly localization results for medical images.

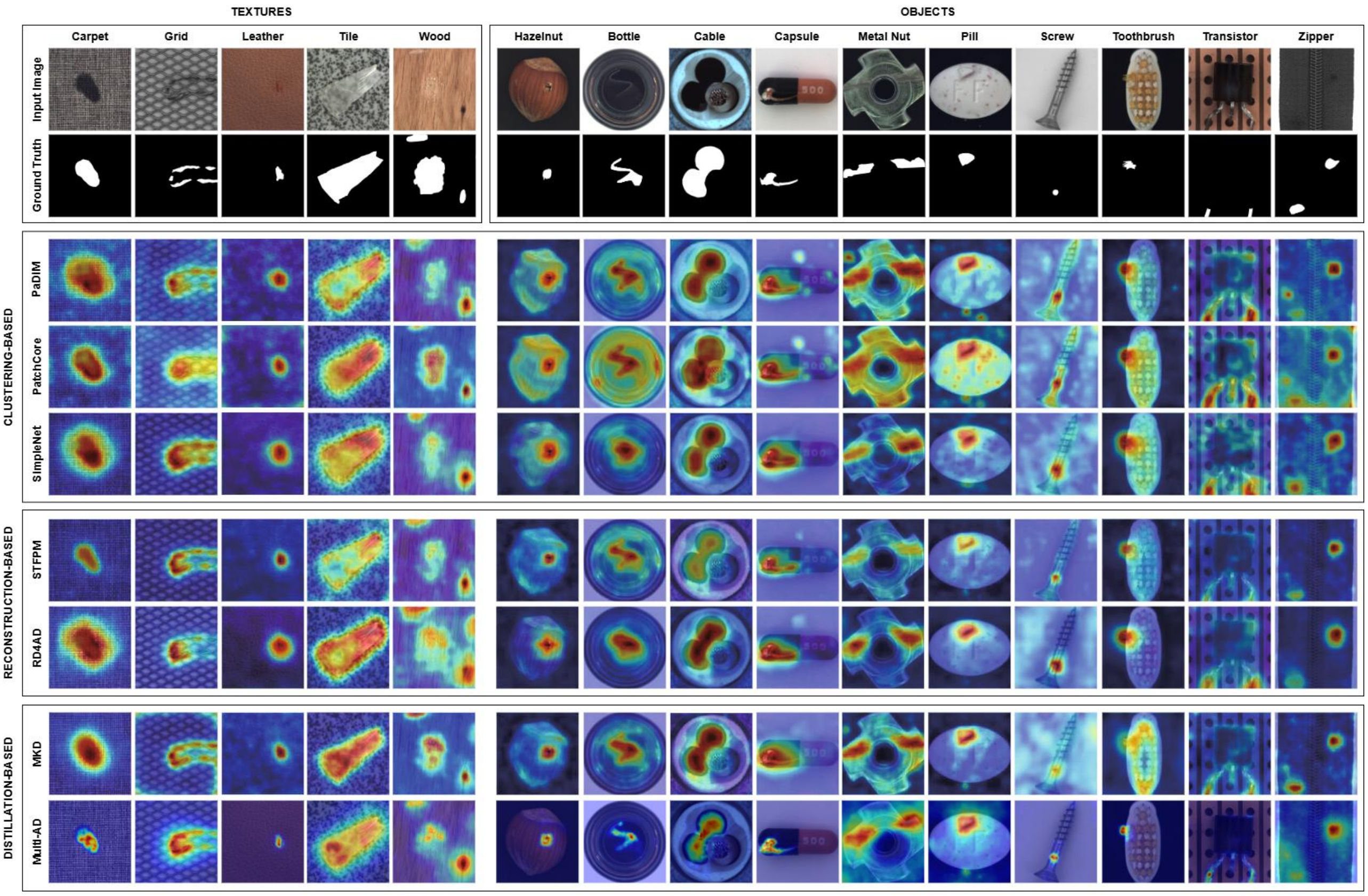

**Figure 7.** Anomaly localization results for industrial images.

b) Industrial

The MVTec AD [44] dataset has been widely used as a benchmark for AD in real-world industrial settings. This dataset comprises 15 categories: 5 texture classes (carpet, grid, leather, tile, and wood) and 10 object classes (hazelnut, bottle, cable, capsule, metal nut, pill, screw, toothbrush, transistor, and zipper), totaling 5,354 images. In addition, it includes 73 types of defects, such as scratches, damage, stains, cracks, deformations, and missing objects.

*4.2 Multi-AD Results*

The performance of the AD model was evaluated using the Area Under the Receiver Operating Characteristic (AUROC) curve, which measures the model's overall ability to distinguish between positive (anomaly) and negative (normal) classes. Higher AUROC values indicate better performance.

Image-level AUROC measures a model's ability to correctly identify whether an entire image is normal or contains anomalies. In contrast, pixel-level AUROC evaluated the model's accuracy in detecting anomalous regions with details at each pixel in an image.

a) Medical

A comparison of AUROC scores for Multi-AD with state-of-the-art models, such as PaDiM [33], PatchCore [30], SimpleNet [45], STFPM [46], RD4AD [47], and MKD [19], is summarized in Table 2 (the best score is shown in bold and the second-best score is underlined). Multi-AD consistently outperformed previous methods, achieving the highest average AUROC scores of 81.4% at the image-level and 97.0% at the pixel-level, respectively, across multiple medical datasets.

Multi-AD outperforms the clustering-based methods, such as PatchCore and SimpleNet, for AD detection on the brain MRI dataset, achieving a pixel-level AUROC of 96.8%. Its performance on the liver CT dataset further highlights its effectiveness, with an image-level AUROC of 62.6% and a pixel-level AUROC of 97.6%. In comparison, the next-best model, STFPM, achieved 61.8% image-level AUROC, whereas SimpleNet achieved 96.8% pixel-level AUROC. Other methods exhibited greater variability and struggled with image-level anomaly detection on the liver CT dataset. Additionally, Multi-AD demonstrates strong performance on the retinal OCT dataset, achieving an image-level AUROC of 91.8%.

Figure 6 shows that Multi-AD consistently produced heatmaps that closely match the ground truth, demonstrating its ability to accurately localize anomalies. Multi-AD sharply localized anomalous regions with minimal false positives (FP), whereas the clustering-based methods (PatchCore, PaDiM, and SimpleNet) produced broader, less focused heatmaps at risk of FP. In contrast, the reconstruction-based methods (STFPM and RD4AD) often failed to capture subtle or spatially complex anomalies.

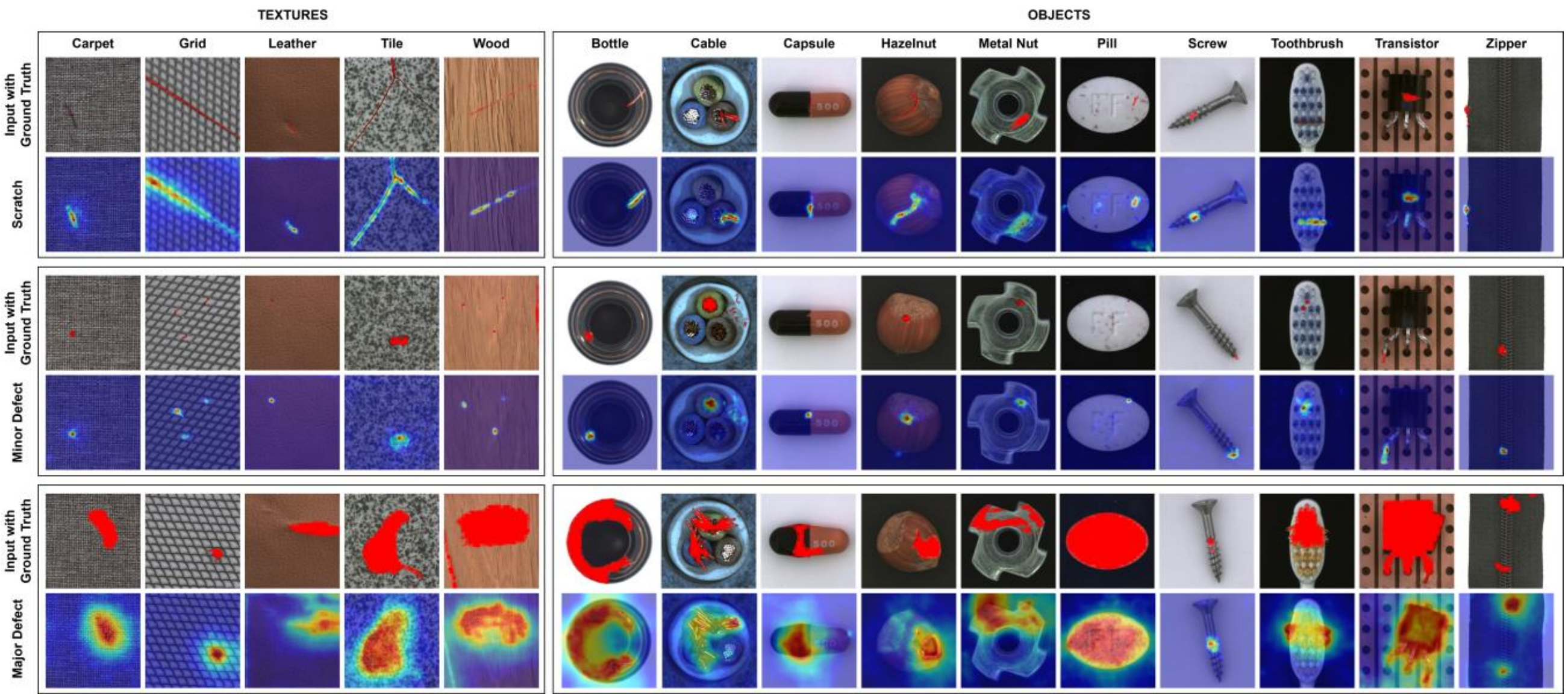

**Figure 8.** Multi-AD results for several cases: scratches, minor defects, and major defects.

In particular, both MKD and Multi-AD belong to the distillation-based category, which leverages knowledge transfer from teacher to student models. Among the two, Multi-AD achieves the best localization performance. This indicates that distillation-based approaches can effectively capture semantic differences in feature representations, making them particularly suitable for medical imaging tasks requiring precise localization and low FP rates.

b) Industrial

A comparison of AUROC scores for Multi-AD against the state-of-the-art models on industrial datasets is summarized in Table 3 (the best score is shown in bold and the second-best score is underlined). Multi-AD achieves the highest average AUROC scores of 99.6% at the image-level and 98.4% at the pixel-level, respectively. Multi-AD consistently outperformed state-of-the-art models, achieving the highest image-level and pixel-level AUROC in several categories. It outperformed clustering-based methods, such as PatchCore, in pixel-level localization, a critical aspect for industrial tasks, including the detection of manufacturing defects or material inconsistencies.

**Table 4.** AUROC (%) results for several ablation experiments.

| Study | Squeeze-Excitation (SE) Block | Discriminator (D) Network | Multi-scale Feature Fusion (MFF) | Medical | | Industrial | |
|---|---|---|---|---|---|---|---|
| | | | | $I_L$ | $P_L$ | $I_L$ | $P_L$ |
| 1 | ✓ | ✓ | ✓ | **81.4** | **97.0** | **99.6** | **98.4** |
| 2 | ✓ | ✓ | - | 78.6 | 91.4 | 93.4 | 94.0 |
| 3 | ✓ | - | ✓ | 79.8 | 93.2 | 96.2 | 95.8 |
| 4 | ✓ | - | - | 77.5 | 90.9 | 92.8 | 91.2 |
| 5 | - | ✓ | ✓ | 78.7 | 95.4 | 95.2 | 96.8 |
| 6 | - | ✓ | - | 72.3 | 89.8 | 91.9 | 92.5 |
| 7 | - | - | ✓ | 73.1 | 91.4 | 93.4 | 95.1 |
| 8 | - | - | - | 70.9 | 86.2 | 89.5 | 88.3 |

Figure 7 shows that Multi-AD consistently produces precise, highly localized heatmaps that align well with ground-truth anomalies across all datasets. For challenging cases, such as bottles and capsules, Multi-AD accurately highlights anomalies with sharp and well-defined heatmaps while avoiding FP in normal regions. It effectively distinguished anomalies from the background on the grid and leather, whereas the clustering-based methods (PaDiM, PatchCore, and SimpleNet) often produced broader, less focused detections. Similarly, Multi-AD demonstrated excellent precision for the pill and zipper by tightly localizing anomalies. Additionally, Multi-AD generated highly specific heatmaps for datasets such as wood and toothbrushes, whereas clustering-based methods struggled with background noise and reduced localization accuracy. In contrast, the reconstruction-based methods (STFPM and RD4AD) showed more diffuse heatmaps, which could lead to over-detection.

In particular, MKD and Multi-AD are distillation-based methods that produce clearer, more concentrated heatmaps and demonstrate superior feature transfer and localization, particularly in texture and object categories. This is particularly evident for complex textures, such as tiles and leather, and for small-object anomalies, such as pills and screws, where distillation-based models outperform clustering and reconstruction approaches.

Figure 8 illustrates the reliability of Multi-AD in localizing various types of industrial anomalies, including scratches, minor defects, and major defects. These specific cases were chosen because they represent the various challenges of industrial quality control. Scratches, for example, are common surface defects that, although often subtle, can cause significant quality issues in manufactured products. Minor defects are equally important because even small imperfections can compromise the functionality or safety of industrial components, especially in precision manufacturing. Finally, major defects are included to assess the model's ability to detect more obvious yet diverse anomalies without compromising accuracy. Multi-AD demonstrated robustness and flexibility by consistently maintaining high detection accuracy across a range of defect types and sizes in industrial environments, thereby improving quality control and operational efficiency.

### *4.3 Ablation Study*

Compared with the original WideResNet architecture and the ablation configuration without the *D* network, Multi-AD significantly improved the model’s performance in detecting subtle and complex anomalies. The SE block provided channel-wise attention, enabling the network to focus on the most significant features of AD. This was particularly relevant in medical imaging, where anomalies could be very small or obscured by complex background patterns, and in industrial imaging, where defects often varied in size and shape. The SE block enabled the model to prioritize the most informative features by recalibrating feature responses, thereby improving anomaly localization and detection accuracy. The network's generalization ability was also enhanced by the SE block, as the model adapted more effectively to different datasets, enabling robust performance across medical and industrial domains, as summarized in Table 4. For example, adding the SE block alone (Study 4) improved both image- and pixel-level AUROC compared to the baseline (Study 8), with medical and industrial pixel-level AUROC increasing from 86.2% to 90.9% and 88.3% to 91.2%, respectively.

The *D* network served as an adversarial component to enhance the AD model's capabilities. By distinguishing between normal and abnormal patterns, the *D* network sharpened the model’s ability to classify subtle anomalies, thereby reducing the FP rate correctly. Models without the *D* network in the original and modified WideResNet architectures showed performance degradation, especially in complex anomaly scenarios. This is evident in the comparison between Study 1 and Study 3, where removing the *D* network reduced medical and industrial pixel-level AUROCs from 97.0% to 93.2% and from 98.4% to 95.8%, respectively. Meanwhile, the SE-enhanced WideResNet outperformed the original WideResNet without the *D* network.

In addition, the MFF module significantly enhances the model's ability to localize anomalies across multiple semantic levels, particularly by improving pixel-level detection accuracy through the integration of contextual information from multiple depths. This is supported by the difference between Study 1 and Study 2, in which MFF increased pixel-level AUROC from 91.4% to 97.0% on medical datasets and from 94.0% to 98.4% on industrial datasets.

AD tasks in medical and industrial imaging differ significantly in feature complexity, anomaly types, and data variability. The SE block and the discriminator allowed the model to adapt to these variations, ensuring robust performance in both domains. This cross-domain robustness is a crucial feature for practical applications in real-world scenarios, where models are often required to operate on diverse datasets without requiring extensive retraining.

## 5. Discussion

Although robust in their respective frameworks, other AD methods often struggle with precise anomaly localization, particularly in complex imaging datasets commonly found in medical and industrial applications. Clustering-based methods employ innovative feature extraction and matching techniques, but sometimes produce broader, less focused heatmaps, which can increase the FP rate. This is particularly problematic in medical imaging, where precise delineation of pathological regions is critical for accurate diagnosis. Broader heatmaps may include non-anomalous regions, thereby misleading the diagnostic process. In contrast, Multi-AD's *D* networks and SE blocks enable sharper, more focused anomaly localization, which is critical for reducing diagnostic errors.

Similarly, reconstruction-based methods, such as STFPM, which combine spatial and temporal features for AD, are limited in their ability to effectively localize anomalies in highly textured or complex industrial backgrounds. They perform poorly on subtle anomalies or blend seamlessly into the background. Furthermore, RD4AD, although designed to be robust across a wide range of settings, does not consistently account for the variability and complexity of anomalies. With its enhanced feature recalibration and discrimination capabilities, Multi-AD significantly outperforms these models, maintaining high sensitivity and specificity even under challenging conditions. Distillation-based methods, such as MKD, also demonstrate strong performance by leveraging teacher–student knowledge transfer; however, MKD lacks adversarial refinement and advanced attention mechanisms. This can lead to more diffuse anomaly maps and slightly reduced localization precision relative to Multi-AD, particularly for small or ambiguous defects.

In industrial applications, Multi-AD's ability to achieve excellent image-level AUROC scores in multiple categories demonstrates its superiority in handling a wide range of defect types. This precision is beneficial for maintaining high-quality standards and critical for ensuring safety and efficiency in manufacturing processes. Interestingly, several MVTec AD texture and object categories, such as leather, wood, hazelnut, and capsules, exhibit structural patterns and anomalies that resemble those in medical images, making them useful for evaluating cross-domain generalization and low-contrast anomaly detection.

While other methods have their advantages and specific use cases, Multi-AD's comprehensive approach and enhanced capabilities allow it to consistently outperform these methods across a spectrum of challenging medical and industrial datasets. This comparative advantage results from enhanced backbone integration with KD and *D* networks, which improved the performance and reliability of the overall AD system.

However, integrating the MFF module and adversarial *D* network increases computational complexity, potentially affecting real-time implementation in industrial pipelines or resource-constrained clinical settings. Addressing these limitations is a crucial direction for future work to enhance the model's robustness, scalability, and adaptability across a broader range of practical settings.

## 6. Conclusion

This work presented Multi-AD, an unsupervised CNN-based model that effectively addressed AD challenges across medical and industrial domains. With the incorporation of KD and SE blocks, our approach enhanced feature extraction and captured subtle differences between normal and abnormal data. Adding a $D$ network further enhanced the model's ability to distinguish anomalies, yielding superior performance compared to state-of-the-art models. Through comprehensive experiments, Multi-AD demonstrated strong generalization across diverse datasets, proving its capabilities in real-world tasks such as early disease diagnosis and industrial defect detection. Future work could involve exploring additional optimization strategies, such as integrating more sophisticated attention mechanisms or refining adversarial learning approaches, to enhance the model's scalability and real-time performance across industrial and medical settings. Furthermore, future research may expand the model's evaluation to a broader range of medical imaging data across multiple anatomical regions and modalities, thereby further assessing its generalizability and clinical applicability.

## Acknowledgments

This work was supported in part by the JST GTIE GAP Fund Program, Grant Number GTIE2024_EX10, Japan.

## References

[1] J. Bao, H. Sun, H. Deng, Y. He, Z. Zhang, X. Li, BMAD: Benchmarks for medical anomaly detection, 2023, arXiv preprint, arXiv:2306.11876.
[2] G. Litjens, T. Kooi, B.E. Bejnordi, A.A.A. Setio, F. Ciompi, M. Ghafoorian, J.A.W.M. van der Laak, B. van Ginneken, C.I. Sánchez, A survey on deep learning in medical image analysis, Med Image Anal 42 (2017) 60–88.
[3] G. Xie, J. Wang, J. Liu, J. Lyu, Y. Liu, C. Wang, F. Zheng, Y. Jin, IM-IAD: Industrial image anomaly detection benchmark in manufacturing, IEEE Trans Cybern 54 (2024) 2720–2733.
[4] Y. Guo, M. Jiang, Q. Huang, Y. Cheng, J. Gong, MLDFR: A multilevel features restoration method based on damaged images for anomaly detection and localization, IEEE Trans Industr Inform 20 (2024) 2477–2486.
[5] J. Zhang, Z. Yang, Y. Song, DC-AD: A divide-and-conquer method for few-shot anomaly detection, Pattern Recognit 162 (2025) 111360.
[6] X. Xia, X. Pan, N. Li, X. He, L. Ma, X. Zhang, N. Ding, GAN-based anomaly detection: A review, Neurocomputing 493 (2022) 497–535.

[7] J. Guo, S. Lu, L. Jia, W. Zhang, H. Li, Encoder-decoder contrast for unsupervised anomaly detection in medical images, IEEE Trans Med Imaging 43 (2024) 1102–1112.
[8] Y. Chen, H. Zhang, Y. Wang, Y. Yang, X. Zhou, Q.M.J. Wu, MAMA Net: Multi-scale attention memory autoencoder network for anomaly detection, IEEE Trans Med Imaging 40 (2021) 1032–1041.
[9] N. Madan, N.-C. Ristea, R.T. Ionescu, K. Nasrollahi, F.S. Khan, T.B. Moeslund, M. Shah, Self-supervised masked convolutional transformer block for anomaly detection, IEEE Trans Pattern Anal Mach Intell (2023) 1–18.
[10] F. Haghighi, M. R. H. Taher, M. B. Gotway, J. Liang, Self-supervised learning for medical image analysis: Discriminative, restorative, or adversarial?, Med Image Anal 94 (2024), 103086.
[11] H. Zhao, Y. Li, N. He, K. Ma, L. Fang, H. Li, Y. Zheng, Anomaly detection for medical images using self-supervised and translation-consistent features, IEEE Trans Med Imaging 40 (2021) 3641–3651.
[12] M. Pei, N. Liu, B. Zhao, H. Sun, Self-supervised learning for industrial image anomaly detection by simulating anomalous samples, Int J Comp Intell Syst 16 (2023) 1–15.
[13] D. Cohen Hochberg, H. Greenspan, R. Giryes, A self supervised StyleGAN for image annotation and classification with extremely limited labels, IEEE Trans Med Imaging 41 (2022) 3509–3519.
[14] S. Yoa, S. Lee, C. Kim, H.J. Kim, Self-supervised learning for anomaly detection with dynamic local augmentation, IEEE Access 9 (2021) 147201–147211.
[15] J. Hao, K. Huang, C. Chen, J. Mao, Dual-student knowledge distillation for visual anomaly detection, Complex and Intelligent Systems 10 (2024) 4853–4865.
[16] D. Qin, J.J. Bu, Z. Liu, X. Shen, S. Zhou, J.J. Gu, Z.H. Wang, L. Wu, H.F. Dai, Efficient medical image segmentation based on knowledge distillation, IEEE Trans Med Imaging 40 (2021) 3820–3831.
[17] Z. Han, J. Wan, Y. Li, G. Li, A refined knowledge distillation network for unsupervised anomaly detection, Electronics, 13 (2024) 4793.
[18] G. Hinton, O. Vinyals, J. Dean, Distilling the knowledge in a neural network, 2015, arXiv preprint arXiv:1503.02531.
[19] M. Salehi, N. Sadjadi, S. Baselizadeh, M.H. Rohban, H.R. Rabiee, Multiresolution knowledge distillation for anomaly detection, in: Proceedings of the IEEE Computer Society Conference on Computer Vision and Pattern Recognition, 2020, pp.14897–14907.
[20] X. Wang, Y. Wang, Z. Pan, G. Wang, Unsupervised anomaly detection and localization via bidirectional knowledge distillation, Neural Comput Appl 36 (2024) 18499–18514.
[21] Y. Cao, X. Xu, Z. Liu, W. Shen, Collaborative discrepancy optimization for reliable image anomaly localization, IEEE Trans Industr Inform 19 (2023) 10674–10683.
[22] M. Yang, J. Liu, Z. Yang, Z. Wu, SLSG: Industrial image anomaly detection with improved feature embeddings and one-class classification, Pattern Recognit 156 (2024) 110862.
[23] Q. Wan, L. Gao, X. Li, L. Wen, Industrial image anomaly localization based on Gaussian clustering of pretrained feature, IEEE Trans Industr Inform 69 (2022) 6182–6192.
[24] C. Han, L. Rundo, K. Murao, T. Noguchi, Y. Shimahara, Z.Á. Milacski, S. Koshino, E. Sala, H. Nakayama, S. Satoh, MADGAN: Unsupervised medical anomaly detection GAN using multiple adjacent brain MRI slice reconstruction, BMC Bioinformatics 22 (2021) 1–20.
[25] T. Schlegl, P. Seeböck, S. M. Waldstein, G. Langs, f-AnoGAN: Fast unsupervised anomaly detection with generative adversarial networks, Med Image Anal 54 (2019) 30–44.
[26] R. Zhang, H. Wang, M. Feng, Y. Liu, G. Yang, Dual-constraint autoencoder and adaptive weighted similarity spatial attention for unsupervised anomaly detection, IEEE Trans Industr Inform 20 (2024) 9393–9403.
[27] N. Shvetsova, B. Bakker, I. Fedulova, H. Schulz, D. V. Dylov, Anomaly detection in medical imaging with deep perceptual autoencoders, IEEE Access 9 (2021) 118571–118583.
[28] H. H. Nguyen, C.N. Nguyen, X.T. Dao, Q.T. Duong, D.P.T. Kim, M.-T. Pham, Variational autoencoder for anomaly detection: A comparative study, 2024, arXiv preprint arXiv:2408.13561.
[29] Q. Zhao, Y. Wang, B. Wang, J. Lin, S. Yan, W. Song, A. Liotta, J. Yu, S. Gao, W. Zhang, MSC-AD: A multiscene unsupervised anomaly detection dataset for small defect detection of casting surface, IEEE Trans Industr Inform 20 (2024) 6041–6052.
[30] K. Roth, L. Pemula, J. Zepeda, B. Scholkopf, T. Brox, P. Gehler, Towards total recall in industrial anomaly detection, in: Proceedings of the IEEE Computer Society Conference on Computer Vision and Pattern Recognition, IEEE Computer Society, 2022, pp. 14298–14308.
[31] Q. Wan, L. Gao, X. Li, L. Wen, Unsupervised image anomaly detection and segmentation based on pretrained feature mapping, IEEE Trans Industr Inform 19 (2023) 2330–2339.
[32] J. Yu, Y. Zheng, X. Wang, W. Li, Y. Wu, R. Zhao, L. Wu, FastFlow: Unsupervised anomaly detection and localization via 2D normalizing flows, 2021, arXiv preprint arXiv:2111.07677.
[33] T. Defard, A. Setkov, A. Loesch, R. Audigier, PaDiM: A patch distribution modeling framework for anomaly detection and localization, Lecture Notes in Computer Science (Including Subseries Lecture Notes in Artificial Intelligence and Lecture Notes in Bioinformatics) 12664 LNCS (2021) 475–489.
[34] X. Tao, D. Zhang, W. Ma, Z. Hou, Z.F. Lu, C. Adak, Unsupervised anomaly detection for surface defects with dual-siamese network, IEEE Trans Industr Inform 18 (2022) 7707–7717.
[35] Z. Hu, X. Zeng, Y. Li, Z. Yin, E. Meng, Z. Wei, L. Zhu, Z. Wang, MSAttnFlow: Normalizing flow for unsupervised

anomaly detection with multi-scale attention, Pattern Recognit 161 (2025) 111220.
[36] C. Zhang, W. Dai, V. Isoni, A. Sourin, Automated anomaly detection for surface defects by dual generative networks with limited training data, IEEE Trans Industr Inform 20 (2024) 421–431.
[37] H. Yao, Y. Cao, W. Luo, W. Zhang, W. Yu, W. Shen, Prior normality prompt transformer for multiclass industrial image anomaly detection, IEEE Trans Industr Inform (2024).
[38] J. Jiang, J. Zhu, M. Bilal, Y. Cui, N. Kumar, R. Dou, F. Su, X. Xu, Masked swin transformer Unet for industrial anomaly detection, IEEE Trans Industr Inform 19 (2023) 2200–2209.
[39] S. Zagoruyko, N. Komodakis, Wide residual networks, in: Proceedings of the British Machine Vision Conference, BMVC, 2016, pp.87.1-87.12.
[40] J. Hu, L. Shen, S. Albanie, G. Sun, E. Wu, Squeeze-and-Excitation networks, IEEE Trans Pattern Anal Mach Intell 42 (2017) 2011–2023.
[41] U. Baid, S. Ghodasara, S. Mohan, M. Bilello, E. Calabrese, E. Colak, K. Farahani, J. Kalpathy-Cramer, F.C. Kitamura, S. Pati, L.M. Prevedello, J.D. Rudie, C. Sako, R.T. Shinohara, T. Bergquist, R. Chai, J. Eddy, J. Elliott, W. Reade, T. Schaffter, T. Yu, J. Zheng, A.W. Moawad, L.O. Coelho, O. McDonnell, E. Miller, F.E. Moron, M.C. Oswood, R.Y. Shih, L. Siakallis, Y. Bronstein, J.R. Mason, A.F. Miller, G. Choudhary, A. Agarwal, C.H. Besada, J.J. Derakhshan, M.C. Diogo, D.D. Do-Dai, L. Farage, J.L. Go, M. Hadi, V.B. Hill, M. Iv, D. Joyner, C. Lincoln, E. Lotan, A. Miyakoshi, M. Sanchez-Montano, J. Nath, X. V. Nguyen, M. Nicolas-Jilwan, J.O. Jimenez, K. Ozturk, B.D. Petrovic, C. Shah, L.M. Shah, M. Sharma, O. Simsek, A.K. Singh, S. Soman, V. Statsevych, B.D. Weinberg, R.J. Young, I. Ikuta, A.K. Agarwal, S.C. Cambron, R. Silbergleit, A. Dusoi, A.A. Postma, L. Letourneau-Guillon, G.J.G. Perez-Carrillo, A. Saha, N. Soni, G. Zaharchuk, V.M. Zohrabian, Y. Chen, M.M. Cekic, A. Rahman, J.E. Small, V. Sethi, C. Davatzikos, J. Mongan, C. Hess, S. Cha, J. Villanueva-Meyer, J.B. Freymann, J.S. Kirby, B. Wiestler, P. Crivellaro, R.R. Colen, A. Kotrotsou, D. Marcus, M. Milchenko, A. Nazeri, H. Fathallah-Shaykh, R. Wiest, A. Jakab, M.-A. Weber, A. Mahajan, B. Menze, A.E. Flanders, S. Bakas, The RSNA-ASNR-MICCAI BraTS 2021 benchmark on brain tumor segmentation and radiogenomic classification, 2021, arXiv preprint arXiv:2107.02314.
[42] J. Hu, Y. Chen, Z. Yi, Automated segmentation of macular edema in OCT using deep neural networks, Med Image Anal 55 (2019) 216–227.
[43] P. Bilic, P. Christ, H.B. Li, E. Vorontsov, A. Ben-Cohen, G. Kaissis, A. Szeskin, C. Jacobs, G.E.H. Mamani, G. Chartrand, F. Lohöfer, J.W. Holch, W. Sommer, F. Hofmann, A. Hostettler, N. Lev-Cohain, M. Drozdzal, M.M. Amitai, R. Vivanti, J. Sosna, I. Ezhov, A. Sekuboyina, F. Navarro, F. Kofler, J.C. Paetzold, S. Shit, X. Hu, J. Lipková, M. Rempfler, M. Piraud, J. Kirschke, B. Wiestler, Z. Zhang, C. Hülsemeyer, M. Beetz, F. Ettlinger, M. Antonelli, W. Bae, M. Bellver, L. Bi, H. Chen, G. Chlebus, E.B. Dam, Q. Dou, C.W. Fu, B. Georgescu, X. Giró-i-Nieto, F. Gruen, X. Han, P.A. Heng, J. Hesser, J.H. Moltz, C. Igel, F. Isensee, P. Jäger, F. Jia, K.C. Kaluva, M. Khened, I. Kim, J.H. Kim, S. Kim, S. Kohl, T. Konopczynski, A. Kori, G. Krishnamurthi, F. Li, H. Li, J. Li, X. Li, J. Lowengrub, J. Ma, K. Maier-Hein, K.K. Maninis, H. Meine, D. Merhof, A. Pai, M. Perslev, J. Petersen, J. Pont-Tuset, J. Qi, X. Qi, O. Rippel, K. Roth, I. Sarasua, A. Schenk, Z. Shen, J. Torres, C. Wachinger, C. Wang, L. Weninger, J. Wu, D. Xu, X. Yang, S.C.H. Yu, Y. Yuan, M. Yue, L. Zhang, J. Cardoso, S. Bakas, R. Braren, V. Heinemann, C. Pal, A. Tang, S. Kadoury, L. Soler, B. van Ginneken, H. Greenspan, L. Joskowicz, B. Menze, The Liver Tumor Segmentation Benchmark (LiTS), Med Image Anal 84 (2023) 102680.
[44] P. Bergmann, M. Fauser, D. Sattlegger, C. Steger, MVTec AD-A comprehensive real-world dataset for unsupervised anomaly detection, in: Proceedings of the IEEE Computer Society Conference on Computer Vision and Pattern Recognition, 2019, pp.9584–9592.
[45] Z. Liu, Y. Zhou, Y. Xu, Z. Wang, SimpleNet: A simple network for image anomaly detection and localization, in: Proceedings of the IEEE Computer Society Conference on Computer Vision and Pattern Recognition, 2023, pp.20402–20411.
[46] G. Wang, S. Han, E. Ding, D. Huang, Student-teacher feature pyramid matching for anomaly detection, 2021, arXiv preprint arXiv:2103.04257.
[47] H. Deng, X. Li, Anomaly detection via reverse distillation from one-class embedding, in: Proceedings of the IEEE Computer Society Conference on Computer Vision and Pattern Recognition, IEEE Computer Society, 2022, pp.9727–9736.